\documentclass[conference]{IEEEtran}

\usepackage[utf8]{inputenc}
\usepackage{graphicx}
\usepackage{amsmath}
\usepackage{amsfonts}
\usepackage{booktabs}
\usepackage{cite}
\usepackage[
colorlinks=true,
linkcolor=blue,
urlcolor=blue,
citecolor=blue
]{hyperref}

\begin{document}

\title{GeoVision-Enabled Digital Twin for Hybrid Autonomous--Teleoperated Medical Responses}

\author{
\IEEEauthorblockN{
    Parham Kebria\IEEEauthorrefmark{1}, 
     Soheil Sabri\IEEEauthorrefmark{2}, and 
   Laura J Brattain\IEEEauthorrefmark{3}}
\IEEEauthorblockA{
    \IEEEauthorrefmark{1}Electrical and Computer Engineering / North Carolina A\&T State University / Greensboro, USA \\ \IEEEauthorrefmark{2}Urban Digital Twin Lab, School of Modeling Simulation and Training / University of Central Florida / Orlando, USA \\\IEEEauthorrefmark{3}Department of Internal Medicine / College of Medicine, University of Central Florida / Orlando, USA \\
Email: 
pmkebria@ncat.edu, \{Soheil.Sabri, Laura.Brattain\}@ucf.edu}
}


\maketitle

\begin{abstract}
Remote medical response systems are increasingly being deployed to support emergency care in disaster-affected and infrastructure-limited environments. Enabled by GeoVision capabilities, this paper presents a Digital Twin architecture for hybrid autonomous–teleoperated medical response systems. The proposed framework integrates perception and adaptive navigation with a Digital Twin, synchronized in real-time, that mirrors system states, environmental dynamics, patient conditions, and mission objectives. Unlike traditional ground control interfaces, the Digital Twin provides remote clinical and operational users with an intuitive, continuously updated virtual representation of the platform and its operational context, enabling enhanced situational awareness and informed decision-making.
\end{abstract}


\section{Introduction}
Remote and prehospital medical response in disaster-affected environments faces severe constraints in infrastructure, communication, and situational awareness \cite{hamilton2022role,dara2005worldwide}. Existing autonomous or teleoperated platforms are typically siloed, with limited integration of patient-state intelligence and environmental context. Digital Twin (DT) technology provides a unified cyber-physical representation that can couple navigation, environment modeling, and clinical decision support.

This paper introduces a geospatially aware digital twin architecture for hybrid visual, autonomous, and teleoperated medical response applications. Unlike conventional ground control interfaces, the Digital Twin offers remote clinical and operational users a dynamic, continuously synchronized virtual representation of the platform and its surrounding environment, thereby enhancing situational awareness and supporting more informed, timely decision-making. This paper makes the following contributions:
\begin{itemize}
    \item Outlining a GeoVision-enabled Digital Twin architecture for hybrid autonomous--teleoperated medical response, explicitly integrating visual geolocalization, adaptive navigation, and healthcare intelligence;
    \item Defining the control and data-flow mechanisms that support dynamic switching between autonomous assistance and operator-in-the-loop control under degraded communication and localization;
    \item Report promising early simulation-based results that indicate improved mission prioritization and operator situational awareness compared to baseline schemes, and also identify key research directions for further development.
\end{itemize}

\begin{figure}
    \centering
    \includegraphics[width=.9\columnwidth]{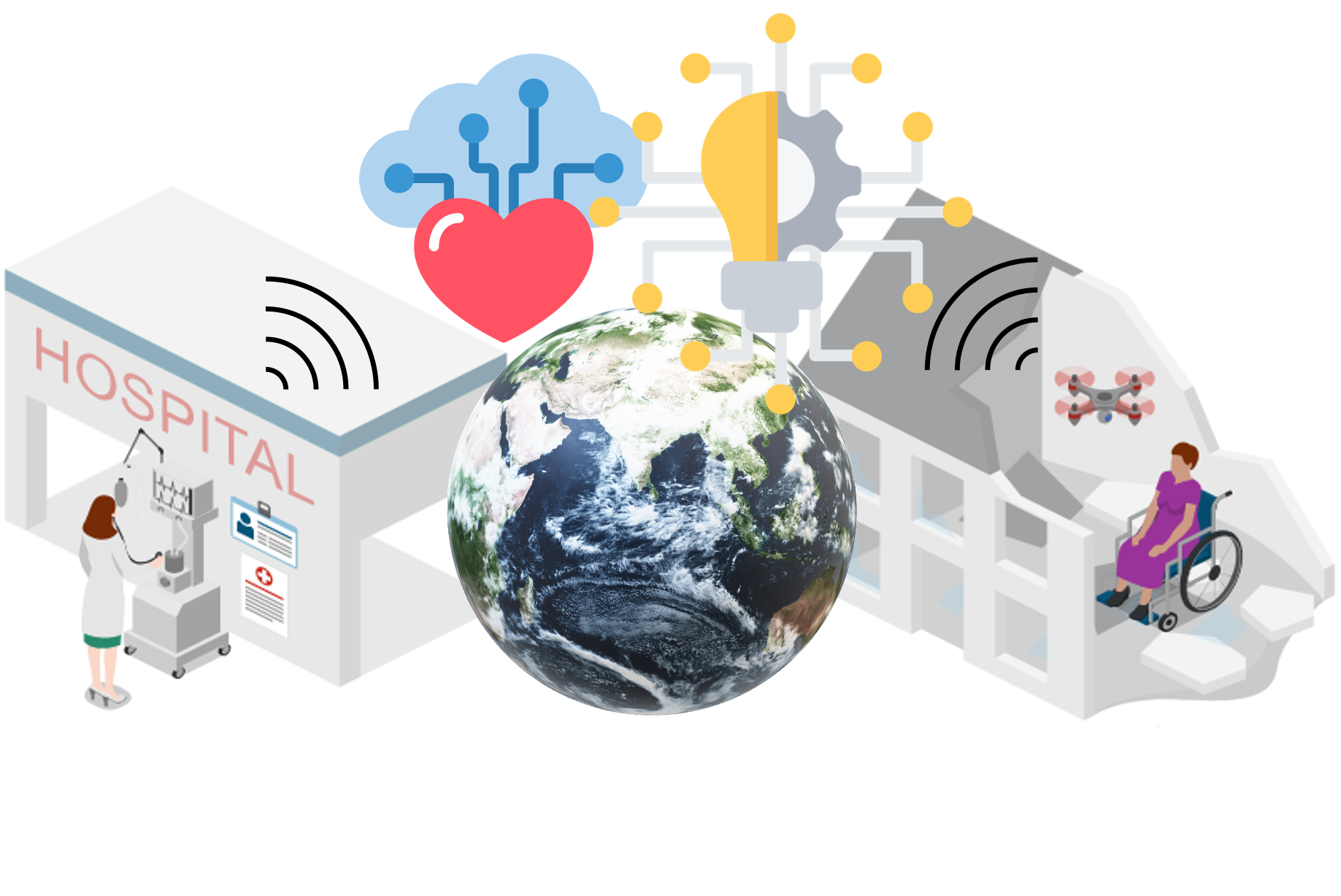}
    \caption{The proposed GeoVision-enabled digital twin framework provides medical services to people in remote and inaccessible areas. Healthcare providers can communicate and monitor the patient's vital signs while supervising the entire situation through the digital twin in real-time.}
    \label{fig:demo}
\end{figure}

\section{Related Work}
Digital Twins (DTs) are increasingly being explored in healthcare and emergency response to integrate heterogeneous data streams and support predictive, risk-aware decision-making \cite{alazab2022digital}. DTs were used for a real-time digital representation of physical robots, allowing operators to visualize remote motions with high precision \cite{khan_development_2023}. This technology facilitates bidirectional communication, providing the real-time monitoring and feedback necessary to tailor therapy programs that would otherwise be challenging to manage remotely \cite{kim_extensive_2022}. Hybrid teleoperation frameworks and robotic medical systems demonstrate the value of combining autonomy with human oversight \cite{laaki2019prototyping,kebria2025hercules}, but often lack a unified, geo-registered world model \cite{lee2022control}. 

Visual geolocalization and geospatial AI (GeoAI) offer robust localization in GPS-degraded environments \cite{thakkar_geopose-enabled_2025}, while AI-driven triage and risk prediction provide patient-centered prioritization. However, prior work rarely couples these elements within a single Digital Twin architecture for remote medical response, which is the focus of this paper.

\section{GeoVision-Enabled Digital Twin Architecture}

\subsection{System Overview}
\begin{figure*}
    \centering
    \includegraphics[width=.9\linewidth]{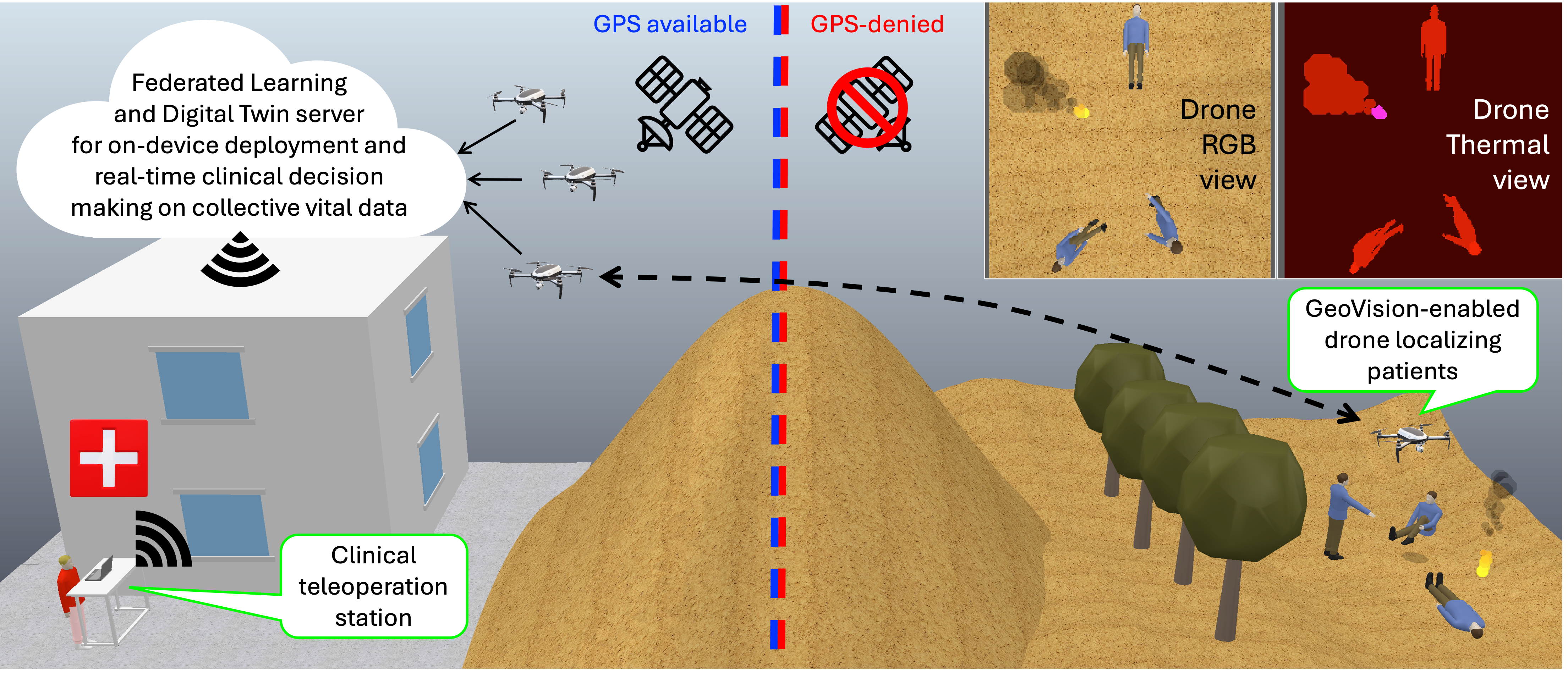}
    \caption{The proposed GeoVision-enabled Digital Twin framework for medical and emergency response applications. GeoVision-enabled drone fleet traverses disaster zones to collect vital data, using Federated Learning to collaboratively update detection on-device models to preserve privacy and bandwidth in communication-limited environments.}
    \label{fig:sysarchitect}
\end{figure*}

The proposed system consists of heterogeneous medical response platforms (e.g., UAVs, UGVs), communication links, and remote clinical/operational users connected through a central Digital Twin (Figure \ref{fig:sysarchitect}). The twin maintains a synchronized representation of the state of the platform, environmental conditions, and patient-related information, serving as the primary interface for clinicians and operators.

Figure \ref{fig:sysarchitect} presents a layered architecture for the proposed GeoVision-enabled Digital Twin framework that incorporates a hybrid autonomous–teleoperated medical response system structured around perception, geolocalization, uncertainty characterization, and risk awareness. Each horizontal layer builds upon the previous one, forming a closed-loop cyber-physical ecosystem.

\subsubsection{Physical Systems Layer} This layer handles real-world sensing, localization, navigation, and mission execution. The perception module fuses LiDAR, camera, and GNSS/IMU data for state estimation and scene understanding. The GeoVision component performs geo-registration and semantic mapping, while Adaptive Navigation manages risk-aware trajectory planning and hybrid control execution. Mission Management coordinates multi-agent task allocation and objective tracking. This layer represents the operational aerial vehicle(s) interacting with the environment.

\subsubsection{GeoVision Bridge} transforms raw spatial and environmental data into structured, geo-registered world models. It maintains 3D scene representations, computes reachability and feasibility constraints, and propagates uncertainty estimates. This layer acts as a structured spatial intelligence engine, enabling the system to understand not just where it is, but what actions are possible and how reliable they are.

\subsubsection{The Digital Twin Layer} is the core cyber-physical interface. It synchronizes platform states and environmental conditions at a certain level of frequency (real-time, near real-time), creating a continuously updated virtual representation. Beyond mirroring, it performs predictive simulation, scenario testing, uncertainty analysis, and risk optimization. It allows “what-if” evaluation before execution and supports multi-objective safety and efficiency trade-offs. This layer transforms the system from reactive telemetry monitoring into predictive operational intelligence.

\subsubsection{The Healthcare Intelligence Layer} integrates medical AI into the mission loop. It performs AI-driven triage, risk prediction, and response prioritization based on injury severity, vital signs, and deterioration probability. Instead of dispatch decisions being purely geographic, they are patient-aware and urgency-weighted. Clinical outputs feed structured reports and decision support tools, ensuring medical intelligence directly influences aerial mission planning.

\subsubsection{Operator and Control Layer} enables human-in-the-loop supervision. Ground control interface visualizes the GeoVision-aligned map and Digital Twin environment. A hybrid control arbiter dynamically balances autonomy and teleoperation depending on mission risk and communication reliability. Remote clinicians interact through intuitive visualizations and live data streams, while the disaster environment module accounts for infrastructure degradation and dynamic hazards. This layer ensures the system remains human-centered, with autonomy acting as decision support rather than replacement.

Together, the architecture forms a human-centered, AI-integrated loop with a digital twin at its core for time-critical predictions and geo-locally informed decision-making in remote healthcare services. 

\subsection{GeoVision Visual Geolocalization Layer}
GeoVision is a multi-stage visual geolocalization subsystem that fuses onboard imagery, inertial sensing, and degraded GNSS signals to estimate platform pose and local environment structure in GPS-challenged settings. It outputs a geo-referenced pose with uncertainty, along with semantic cues such as road segments, access points, and hazards inferred from visual perception. These outputs feed both the navigation stack and the Digital Twin, enabling a geo-registered world model even when traditional localization is unreliable.

\subsection{Navigation and Mission Management}
The navigation layer consumes GeoVision pose estimates and Digital Twin map layers to generate risk-aware trajectories that account for damaged infrastructure, access constraints, and communication conditions. A mission manager coordinates task assignment and route selection based on both navigation feasibility and clinical priorities, enabling dynamic re-planning as new information arrives. When localization uncertainty or communication degradation crosses predefined thresholds, the system transitions between autonomous, assisted, and teleoperated modes.

\subsection{Healthcare Intelligence and Operator Interface}

The Healthcare Intelligence Module forms the core of clinical decision support within this framework. It provides AI-driven triage assessment and patient risk prediction by processing multimodal inputs such as vital signs (e.g., heart rate, oxygen saturation), medical imaging, and environmental indicators. Built upon a pretrained foundation model, the module generates clinically meaningful outputs including severity scores, deterioration probabilities, and intervention recommendations \cite{moor2023foundation, zhou2023foundation}. These predictions are dynamically updated through online learning mechanisms, allowing the model to incorporate newly observed patient trajectories and adapt to context-specific patterns such as region-specific disease prevalence or disaster-induced health risks.
To further enhance robustness, when GPS is available, the module integrates federated continual learning, where distributed agents (e.g., drones, mobile medical units, field hospitals) collaboratively improve the shared model while retaining data locally \cite{kairouz2021advances, li2020federated}. This is particularly critical in healthcare scenarios where data privacy, regulatory compliance, and limited connectivity are key constraints. The combination of foundation models and federated learning not only mitigates the challenge of limited labeled physiological data but also enables personalized and context-aware predictions across diverse populations and environments \cite{rieke2020future}.
These AI-generated outputs are fused with GeoVision-informed travel-time estimates, terrain accessibility, and mission constraints to prioritize interventions and optimize resource allocation. The system can, for example, dynamically rank patients based on both medical urgency and logistical feasibility, ensuring that critical cases receive timely attention even in complex disaster scenarios.

Clinicians and operators interact with the system through a Digital Twin interface that visualizes platform pose, candidate trajectories, and patient locations within a shared geo-registered environment. The integration of continuously learning AI models with a real-time Digital Twin creates a closed-loop intelligent system, where perception, prediction, and decision-making are tightly coupled to support efficient, adaptive, and scalable remote medical response. We will adapt our existing work in fine-tuning foundation models to vital signs \cite{DBLP:conf/bsn/AliMBB24} and to federated learning \cite{DBLP:journals/corr/abs-2502-09744} to build this Healthcare Intelligence Module. 
\section{Methodology and Early Results}

\subsection{Simulation Setup}
To obtain early indications of feasibility, we construct a simplified simulation scenario representing a disaster-affected urban area with multiple patients, degraded GNSS coverage, and intermittent communication. Heterogeneous platforms are modeled with basic kinematics and sensing constraints, while GeoVision is abstracted as a localization module that provides pose estimates with configurable accuracy and outage patterns. We compare three strategies: (i) baseline teleoperation with nominal GPS, (ii) fully autonomous navigation with heuristic mission ordering, and (iii) the proposed GeoVision-enabled Digital Twin framework with triage-aware mission management.

\subsection{Comparative Evaluation Framework}

We evaluate three strategies across the defined performance metrics. Let
$\Pi = \{\pi_1, \pi_2, \pi_3\}$ denote the set of candidate strategies, formally
defined as follows.

\subsubsection*{Strategy $\pi_1$ — Baseline Teleoperation with Nominal GPS}

Strategy $\pi_1$ represents a human-in-the-loop teleoperation baseline in which
an operator directly controls the UAV under nominal GPS availability. The
localization state at time $t$ is given by the GPS-provided position estimate
$\hat{\mathbf{x}}_t^{\text{GPS}} \in \mathbb{R}^3$, with associated covariance
$\boldsymbol{\Sigma}_t^{\text{GPS}}$. Mission ordering follows the operator
discretion, and no onboard autonomy or triage prioritization is applied. The
control policy is:

\begin{equation}
    \pi_1: \quad \mathbf{u}_t = f_{\text{op}}\!\left(\hat{\mathbf{x}}_t^{\text{GPS}},\,
    \mathcal{O}_t\right)
    \label{eq:pi1}
\end{equation}

\noindent where $\mathbf{u}_t$ is the control input at time $t$, $f_{\text{op}}$
denotes the operator control law, and $\mathcal{O}_t$ is the set of observations
available to the operator at time $t$.

\subsubsection*{Strategy $\pi_2$ — Fully Autonomous Navigation with Heuristic Mission Ordering}

Strategy $\pi_2$ replaces the human operator with an onboard autonomous navigation
stack. Localization relies on a self-contained state estimator
$\hat{\mathbf{x}}_t^{\text{auto}}$ independent of GPS. Mission sequencing follows
a heuristic ordering function $\sigma_h$, such as nearest-neighbor or
first-detected priority, applied to the current patient set $\mathcal{P}_t$:

\begin{equation}
    \pi_2: \quad \mathbf{u}_t = f_{\text{auto}}\!\left(\hat{\mathbf{x}}_t^{\text{auto}},\,
    \sigma_h(\mathcal{P}_t)\right)
    \label{eq:pi2}
\end{equation}
where $f_{\text{auto}}$ is the autonomous navigation controller and
$\sigma_h: 2^{\mathcal{P}} \rightarrow \mathcal{P}^*$ maps the patient set to a
heuristically ordered visit sequence.

\subsubsection*{Strategy $\pi_3$ — Proposed GeoVision-Enabled Digital Twin Framework with Triage-Aware Mission Management}

Strategy $\pi_3$ constitutes the proposed framework, integrating a GeoVision
perception module with a Digital Twin representation of the operational
environment and a triage-aware mission planner. Let
$\mathcal{DT}_t$ denote the Digital Twin state at time $t$, encoding terrain,
patient locations, severity scores, and system health. The triage-aware ordering
function $\sigma_{\tau}$ assigns a priority score $v_i$ to each patient
$i \in \mathcal{P}_t$ based on severity $s_i$, estimated time-to-criticality
$\Delta_i$, and UAV accessibility $a_i$:

\begin{equation}
    v_i = g(s_i,\, \Delta_i,\, a_i), \quad \forall\, i \in \mathcal{P}_t
    \label{eq:triage_score}
\end{equation}

\noindent The mission sequence is then determined by:

\begin{equation}
    \sigma_{\tau}(\mathcal{P}_t) = \operatorname*{arg\,sort}_{i \in \mathcal{P}_t}
    \; v_i
    \label{eq:triage_order}
\end{equation}

\noindent The control policy under $\pi_3$ is:

\begin{equation}
    \pi_3: \quad \mathbf{u}_t = f_{\text{GV}}\!\left(\hat{\mathbf{x}}_t^{\text{DT}},\,
    \sigma_{\tau}(\mathcal{P}_t),\, \mathcal{DT}_t\right)
    \label{eq:pi3}
\end{equation}

\noindent where $f_{\text{GV}}$ is the GeoVision-informed navigation controller
and $\hat{\mathbf{x}}_t^{\text{DT}}$ is the Digital Twin-fused localization
estimate, incorporating GPS, inertial, and visual-geometric cues to maintain
positioning integrity under degraded conditions.

\subsubsection*{Strategy Comparison}

The three strategies are evaluated under a common experimental condition set
$\mathcal{E}$, spanning varying levels of GPS degradation $\delta \in [0,1]$
and patient load $|\mathcal{P}|$. For each strategy $\pi_j \in \Pi$ and
condition $e \in \mathcal{E}$, performance is characterized by the metric
vector:

\begin{equation}
    \mathbf{m}(\pi_j, e) = \left[\,\bar{T}_{\text{int}},\; \rho,\;
    R_{\text{fail}}(\delta),\; \bar{W}\,\right]^{\!\top}
    \label{eq:metric_vector}
\end{equation}

\noindent A strategy $\pi_j$ is said to \textit{dominate} $\pi_k$ under condition
$e$ if:

\begin{equation}
    \mathbf{m}(\pi_j, e) \preceq \mathbf{m}(\pi_k, e), \quad \pi_j \neq \pi_k
    \label{eq:dominance}
\end{equation}

\noindent where $\preceq$ denotes component-wise inequality (lower is better for
all metrics), with strict inequality in at least one component.

\subsection{Evaluation Metrics}
We focus on metrics that capture both mission effectiveness and human factors:
average time to first intervention for high-severity patients, number of patients served within a clinically acceptable time window, mission failure or abort rate under localization/communication degradation, and proxy indicators of operator workload derived from task-switching and intervention frequency.

\subsubsection*{Metric 1 — Average Time to First Intervention (High-Severity Patients)}

Let $\mathcal{P}_H \subseteq \mathcal{P}$ denote the set of high-severity patients,
where $|\mathcal{P}_H| = N_H$. For each patient $i \in \mathcal{P}_H$, let
$t_i^{\text{detect}}$ be the time of detection and $t_i^{\text{int}}$ be the time
of first intervention. The average time to first intervention is:

\begin{equation}
    \bar{T}_{\text{int}} = \frac{1}{N_H} \sum_{i \in \mathcal{P}_H}
    \left( t_i^{\text{int}} - t_i^{\text{detect}} \right)
    \label{eq:avg_intervention_time}
\end{equation}

\subsubsection*{Metric 2 — Patients Served Within Clinically Acceptable Time Window}

Let $\tau_c$ denote the clinically acceptable time threshold. The count of patients
served within this window is:

\begin{equation}
    N_{\text{served}} = \sum_{i \in \mathcal{P}}
    \mathbf{1}\!\left[ t_i^{\text{int}} - t_i^{\text{detect}} \leq \tau_c \right]
    \label{eq:patients_served}
\end{equation}

The corresponding service rate is $\rho = N_{\text{served}} / |\mathcal{P}|$.

\subsubsection*{Metric 3 — Mission Failure/Abort Rate Under Degradation}

Let $\mathcal{M}$ be the set of all mission trials conducted under localization or
communication degradation, with $|\mathcal{M}| = M$. Let
$\mathcal{M}_F \subseteq \mathcal{M}$ be the subset of missions that result in
failure or abort. The mission failure rate is:

\begin{equation}
    R_{\text{fail}} = \frac{|\mathcal{M}_F|}{M}
    = \frac{1}{M}\sum_{m=1}^{M} \mathbf{1}[m \in \mathcal{M}_F]
    \label{eq:failure_rate}
\end{equation}

This may further be conditioned on a degradation severity level $\delta \in [0,1]$,
yielding $R_{\text{fail}}(\delta)$ to characterize system robustness as a function
of degradation.

\subsubsection*{Metric 4 — Operator Workload Proxy}

Let $S_k^{(o)}$ denote the $k$-th task performed by operator $o$ over a mission of
duration $T$, drawn from a task-type alphabet $\Sigma$. We define two component
measures:

\noindent\textit{Task-switching rate:}
\begin{equation}
    \lambda_{\text{sw}}^{(o)} = \frac{1}{T}\sum_{k=1}^{K-1}
    \mathbf{1}\!\left[S_{k+1}^{(o)} \neq S_k^{(o)}\right]
    \label{eq:task_switching}
\end{equation}

\noindent\textit{Intervention frequency:}
\begin{equation}
    \lambda_{\text{int}}^{(o)} = \frac{N_{\text{int}}^{(o)}}{T}
    \label{eq:intervention_freq}
\end{equation}

\noindent where $N_{\text{int}}^{(o)}$ is the total number of interventions by
operator $o$. A composite workload proxy is then defined as:

\begin{equation}
    W^{(o)} = \alpha\, \lambda_{\text{sw}}^{(o)} + \beta\, \lambda_{\text{int}}^{(o)}
    \label{eq:workload}
\end{equation}

\noindent where $\alpha, \beta \geq 0$ are weighting coefficients determined
empirically or via a validated workload model (e.g., NASA-TLX mapping). The
system-level workload aggregated across all operators $\mathcal{O}$ is:
\begin{equation}
    \bar{W} = \frac{1}{|\mathcal{O}|}\sum_{o \in \mathcal{O}} W^{(o)}
    \label{eq:avg_workload}
\end{equation}

Together, the metrics $\bar{T}_{\text{int}}$, $N_{\text{served}}$ (or $\rho$),
$R_{\text{fail}}(\delta)$, and $\bar{W}$ jointly capture mission effectiveness
under degraded conditions and operator cognitive burden.

\subsection{Promising Early Results}
Preliminary simulations suggest that the proposed architecture can reduce time-to-intervention for high-severity patients and decrease mission failure rates in GPS-degraded conditions compared to the baselines.
The Digital Twin visualization, driven by GeoVision, appears to support more efficient operator interventions by providing a consolidated view of platform state, environment hazards, and patient priorities.
While these results are based on simplified models and limited scenario diversity, they indicate that tighter coupling between geolocalization, Digital Twin state, and triage intelligence is a promising direction for further development.

\begin{figure}[t]
    \centering
    \includegraphics[width=.99\columnwidth]{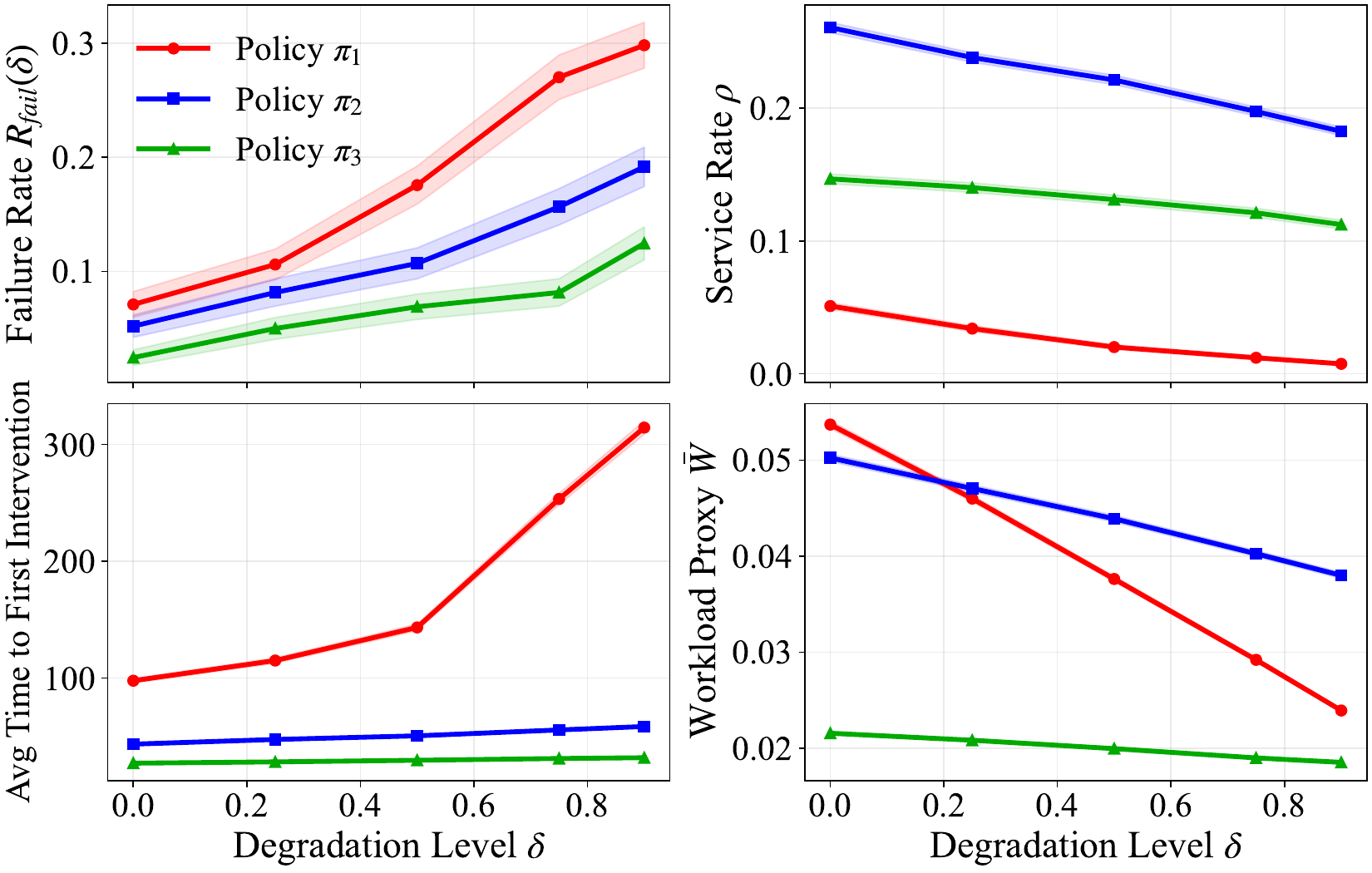}
    \caption{Mean policy performance with 95\% confidence intervals across degradation levels, quantifying both expected outcomes and statistical uncertainty for each core metric.}
    \label{fig:policy_comparison}
\end{figure}
\begin{figure}[ht]
    \centering
    \includegraphics[width=\columnwidth]{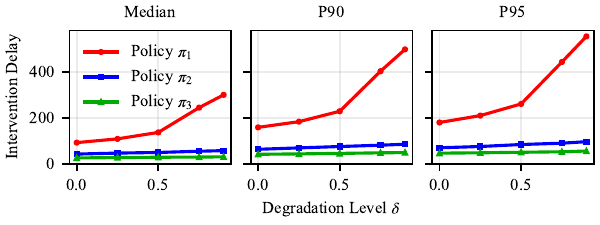}
    \caption{Median, P90, and P95 high-severity intervention delays versus degradation, emphasizing tail-risk behavior and worst-case clinical responsiveness by policy.}
    \label{fig:latency}
\end{figure}
\begin{figure}[ht]
    \centering
    \includegraphics[width=\columnwidth]{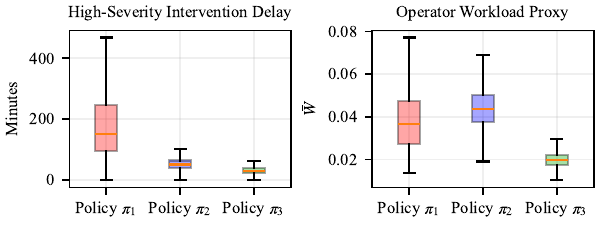}
    \caption{Distributional comparison of high-severity intervention delay and operator workload by policy, showing central tendency and spread beyond mean-only summaries.}
    \label{fig:distribution}
\end{figure}
\begin{figure}[ht]
    \centering
    \includegraphics[width=\columnwidth]{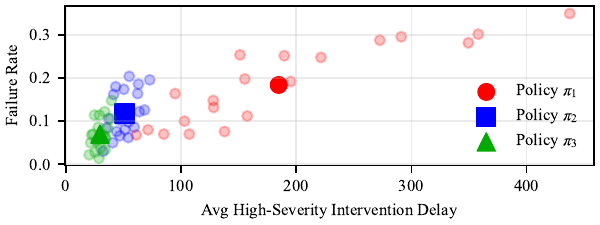}
    \caption{Pareto trade-off between intervention delay and failure rate across operating conditions; larger bubbles indicate higher effective success, illustrating multi-objective policy positioning.}
    \label{fig:tradeoff}
\end{figure}

Table~\ref{tab:policy_summary} and Figure~\ref{fig:policy_comparison} summarize the three-policy comparison across degradation and patient-load conditions. The study used a Monte Carlo framework with stochastic patient-field generation and policy-specific mission execution (teleoperation $\pi_1$, heuristic autonomy $\pi_2$, and triage-aware GeoVision--Digital Twin planning $\pi_3$). A total of 15{,}000 missions were simulated (250 trials per condition over 5 degradation levels, 4 patient-load settings, and 3 policies) with a fixed random seed for reproducibility. Overall, $\pi_3$ provides the strongest balance of robustness, critical-response timeliness, and lower operator burden, while $\pi_2$ primarily emphasizes mission throughput; $\pi_1$ remains the least resilient under degraded conditions.

\begin{table}[t]
\centering
\caption{Policy comparisons across all degradation and patient-load conditions. \textbf{Lower} is better for $\bar{T}_{\mathrm{int}}$, $R_{\mathrm{fail}}$, and $\bar{W}$; \textbf{higher} is better for $\rho$.}
\label{tab:policy_summary}
\resizebox{\columnwidth}{!}{
\begin{tabular}{lccccc}
\hline
Policy & $\bar{T}_{\mathrm{int}}$ (min) & $\rho$ & $R_{\mathrm{fail}}$ & $\bar{W}$ & Mission Time (min) \\
\hline
$\pi_1$ & 182.584         & 0.0247          & 0.1810          & 0.0381          & 392.199 \\
$\pi_2$ & 50.947          & \textbf{0.2209} & 0.1224          & 0.0440          & \textbf{114.439} \\
$\pi_3$ & \textbf{29.546} & 0.1295          & \textbf{0.0664} & \textbf{0.0200} & 140.881 \\
\hline
\end{tabular}}
\end{table}

Across Figures \ref{fig:policy_comparison} to \ref{fig:tradeoff}, the same core result is consistent: Policy $\pi_3$ (GeoVision + DT) is the best balanced option, Policy $\pi_2$ is fastest in aggregate service but with higher operator burden than $\pi_3$, and Policy $\pi_1$ is dominated under degradation. In the CI comparison figure, $\pi_3$ maintains the lowest mean failure rate (about 0.070 vs 0.118 for $\pi_2$ and 0.184 for $\pi_1$), the lowest high-severity intervention delay (about 29.7 vs 51.2 vs 184.8), and the lowest workload proxy (about 0.020 vs 0.0439 vs 0.0381), while $\pi_2$ has the highest mean service rate (about 0.220 vs 0.130 for $\pi_3$ and 0.025 for $\pi_1$), with separation that remains visible across degradation levels. The latency-quantile figure reinforces robustness in the tails: $\pi_3$ shows much tighter and lower intervention-delay tails (P90 about 45.9, P95 about 51.4) than $\pi_2$ (P90 about 76.5, P95 about 85.4), while $\pi_1$ exhibits extreme tail risk (P90 about 362.9, P95 about 432.8). The metric-distribution boxplots show the same pattern distributionally, with $\pi_1$ having the widest and highest delay spread, $\pi_2$ intermediate, and $\pi_3$ the most compact delay distribution, and with workload distributions indicating $\pi_3$ as the lowest-center, most stable workload regime. Finally, the Pareto tradeoff figure (delay vs failure) places $\pi_3$ closest to the lower-left region (better on both objectives), $\pi_2$ as a speed-oriented but less robust compromise, and $\pi_1$ far from the Pareto-efficient frontier, confirming that under this simulation setup $\pi_3$ delivers the strongest reliability-responsiveness compromise.

\section{Conclusions and Future Work}
This paper presented a GeoVision-enabled Digital Twin architecture for hybrid autonomous--teleoperated medical response, emphasizing the integration of visual geolocalization, navigation, and healthcare intelligence. Promising early simulation-based results suggest that this approach can improve mission prioritization and situational awareness in degraded environments, motivating further research toward full-scale implementation and validation.

The current study is limited by its use of abstracted GeoVision models, simplified clinical processes, and a constrained set of simulated disaster scenarios. 
Future work will incorporate higher-fidelity visual geolocalization, more realistic communication and failure models, fine-tuned foundation models with federated learning, integration with existing robotic platforms, and adaptation to practical clinical workflows. We also plan to investigate human-in-the-loop studies to quantify the efficacy of the proposed Digital Twin interface and GeoVision visualization on operator performance and trust.

\bibliographystyle{IEEEtran}
\bibliography{references}

\end{document}